\documentclass[runningheads]{llncs}
\usepackage{graphicx}
\usepackage[caption=false]{subfig}
\usepackage[]{algorithm2e}

\begin{document}
\title{Multi-modal Sentiment Analysis using Super Characters Method on Low-power CNN Accelerator Device}
\titlerunning{Multi-modal Sentiment Analysis using Super Characters}
%
\author{Baohua Sun\inst{1} \and
Lin Yang\inst{1} \and
Hao Sha\inst{1} \and
Michael Lin\inst{1}}
\authorrunning{B. Sun et al.}
%
\institute{Gyrfalcon Technology Inc., 
1900　McCarthy Blvd Suite 208, Milpitas, CA, 95035, US
\email{baohua.sun@gyrfalcontech.com}\\
}
%
%
%
\maketitle              
\begin{abstract}
Recent years NLP research has witnessed the record-breaking accuracy improvement by DNN models. However, power consumption is one of the practical concerns for deploying NLP systems. Most of the current state-of-the-art algorithms are implemented on GPUs, which is not power-efficient and the deployment cost is also very high. On the other hand, CNN Domain Specific Accelerator (CNN-DSA) has been in mass production providing low-power and low cost computation power. In this paper, we will implement the Super Characters method on the CNN-DSA. In addition, we modify the Super Characters method to utilize the multi-modal data, i.e. text plus tabular data in the CL-Aff shared task.

\end{abstract}
\section{Introduction}

The need to classify sentiment based on the multi-modal input arises in many different problems in customer related marketing fields. Super Characters \cite{sun2018super} is a two-step method for sentiment analysis. It first converts text into images; then feeds the images into CNN models to classify the sentiment. Sentiment classification performance on large text contents from customer online comments shows that the Super Character method is superior to other existing methods. The Super Characters method also shows that the pretrained models on a larger dataset help improve accuracy by finetuning the CNN model on a smaller dataset. Compared with from-scratch trained Super Characters model, the finetuned one improves the accuracy from 95.7\% to 97.8\% on the well-known Chinese dataset of Fudan Corpus. Squared English Word (SEW)~\cite{sun2019squared} is an extension of the Super Characters method into Latin Languages. With the wide availability of low-power CNN accelerator chips~\cite{sun2018ultra}~\cite{sun2018mram}, Super Characters method has the great potential to be deployed in large scale by saving power and fast inference speed. In addition, it is easy to deploy as well. The recent work also extend its applications to chatbot~\cite{sun2019superchat}, image captioning~\cite{sun2019supercaptioning}, and also tabular data machine learning~\cite{sun2019supertml}.

The CL-AFF Shared Task\cite{CL-AFF} is part of the Affective Content Analysis workshop at AAAI 2020. It builds upon the OffMyChest dataset\cite{asai2018happydb}, which contains 12,860 samples of training data and 5,000 samples of testing data. Each sample is a multi-modal input containing both text and tabular data. The text input is an English sentence from Reddit. The tabular data is the corresponding log information for each sentence, like wordcount, created utc time and etc. And each sample has six sets of binary classification labels, EmotionDisclosure?(Yes$|$No), InformationDisclosure?(Yes$|$No), Support?(Yes$|$No), EmmotionSupport?(Yes$|$No), InformationSupport?(Yes$|$No), GeneralSupport?(Yes$|$No). In this paper, we will apply Super Characters on this data set to classify the muti-modal input.

\section{Super Characters for Multi-modal Sentiment Analysis and Low-Power Hardware Solution}

For multi-modal sentiment analysis, we can simply split the image into two parts. One for the text input, and the other for the tabular data. Such that both can be embedded into the Super Characters image. The CNN accelerator chip comes together with a Model Development Kit (MDK) for CNN model training, which feeds the two-dimensional Super Characters images into MDK and then obtain the fixed point model. Then, using the Software Development Kit (SDK) to load the model into the chip and send command to the CNN accelerator chip, such as to read an image, or to forward pass the image through the network to get the inference result. The advantage of using the CNN accelerator is low-power, it consumes only 300mw for an input of size 3x224x224 RGB image at the speed of 140fps. Compared with other models using GPU or FPGA, this solution implement the heavy-lifting DNN computations in the CNN accelerator chip, and the host computer is only responsible for memory read/write to generate the designed Super Character image. This has shown good result on system implementations for NLP applications~\cite{sun2019NLPdemo}.

\section{Experiments}
\subsection{Data Exploration}
The training data set has 12,860 samples with 16 columns. The first ten columns are attributes, including sentenceid, author, nchar, created\_utc, score, subreddit, label, full\_text, wordcount, and id. And the other six columns are labels for each of the tasks of Emotion\_disclosure, Information\_disclosure, Support, Emmotion\_support, Information\_support, and General\_support. Each task is a binary classification problem based on the ten attributes. So there will be 60 models to be trained for a 10-fold validation. The test data set has 5000 samples with only the ten columns of attributes. The system run will give labels on these test samples based on the 10-fold training.

For the training data, unique ids are 3634 compared to the whole training 12,860. While for the testing data, this number is only 2443 compared to the whole testing dataset 5000, meaning some of the records may come from the same discussion thread. And the unique authors are 7556 for training, and 3769 for testing, which means some of the authors are active that they may published more than one comments.

Based on this, we have considered to include author names in the multi-modal model as well, since a comment may be biased by the personality of its author. The maximum length of an author's name is 20 charactors, if SEW~\cite{sun2019squared} is to be used to project the names onto a two-dimensional embedding. On the other hand, the nchar which indicates the number of characters for the full\_text has a maximum value of 9993, and the maximum wordcount is 481. The column ``label" has 37 unique values, which are different combinations of strings like ``husband", ``wife", ``boyfriend", ``girlfriend", and their abbreviations like ``bf",``gf". The column ``subreddit" is a categorical attribute with values in (``offmychest", ``CasualConversation"). After converting the Unix time in the column of ``created\_utc", we found that the records are generated from 2017 to 2018. The column score has integers ranging from -44 to 1838 with 251 unique values.

\subsection{Design SuperCharacters Image}
\begin{figure}
\centering
\includegraphics[width=0.8\textwidth]{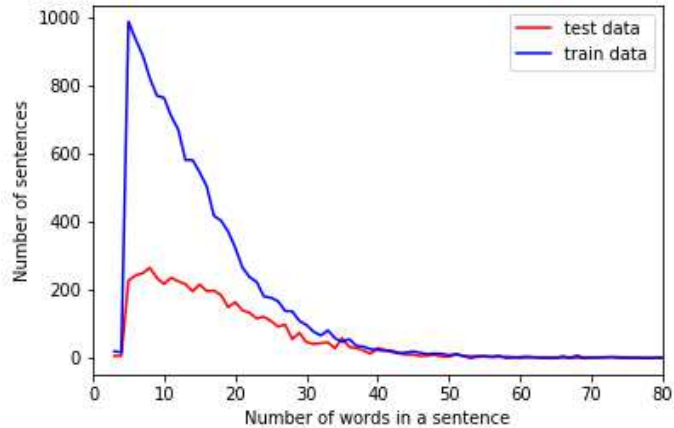}
\caption{Histogram of sentence length counted by number of words.} \label{Fig_TrainData_histogram}
\end{figure}

The sentence length distribution is given in Figure \ref{Fig_TrainData_histogram}. The layout design for the full\_text will be based on this. Since we present the English words using SEW~\cite{sun2019squared} method, the size of each English word on the SuperCharacters image should better be calculated by (224/N)*(224/N) if the whole image is set to 224x224. Here N is an integer. The dimension is set to 224x224 because of the chip specification.

\begin{figure}
\centering
\subfloat[Design Option One: 7 words per row, max 7 rows, with only full\_text information embedded in the image. Example: full\_text=``If it were me, and I cared about a person, I would absolutely read their book to show support, but I can also understand struggling to get stalted on/through something I have absolutely no interest in.". \label{DesignOptionOne}]{
\includegraphics[width=.5\textwidth]{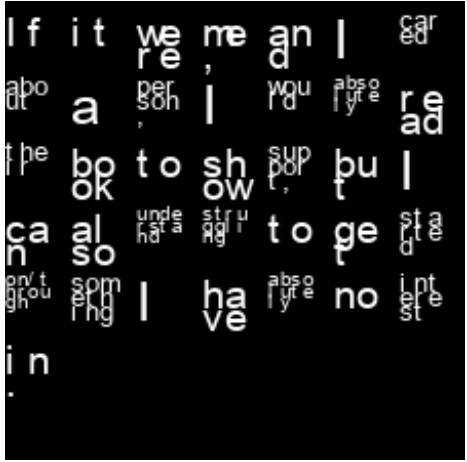}
}~~~~~~
\subfloat[Design Option Two: 8 words per row, max 5 rows, with all attributes except id and sentenceid. Example: author=``Laseyguy", wordcount=32, created\_time=``2018-07-26 19:49:22", subreddit=``offmychest", score=2426, nchar=102, label=``husband", and full\_text=``I think it's safe to say that we've all been there - that realization that this isn't what we signed up for - and that life will never be the same again".\label{DesignOptionTwo}]{
\includegraphics[width=.5\textwidth]{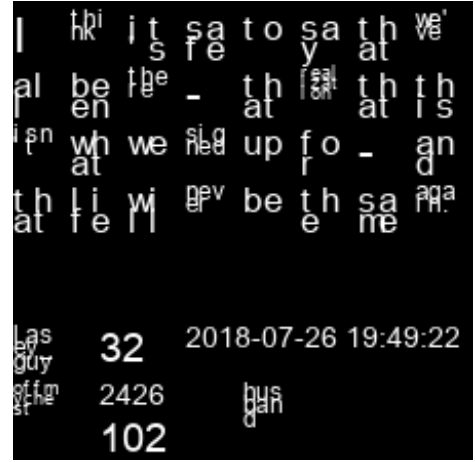}
}\\
\subfloat[Design Option Three: 7 words per row, max 6 rows, with four attributes subreddit, wordcount, score, label, and full\_text. Example: subreddit=``CasualConversation", wordcount=37, score=2, label=``girlfriend", and full\_text=``The best thing that you can do is keep at it don't give up hope, and try not to lose sight of what is important - you health and the health of your loved ones (and kity)!".\label{DesignOptionThree}]{
\includegraphics[width=.5\textwidth]{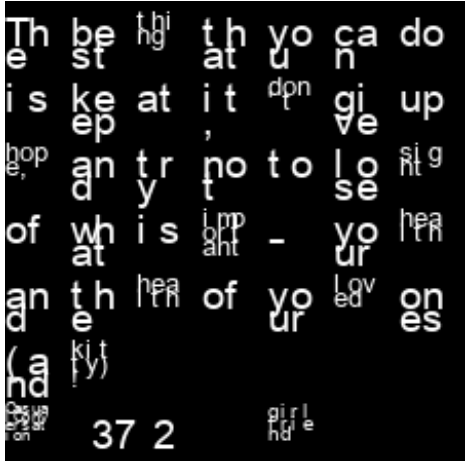}
}~~~~~~
\subfloat[Design Option Four: Same design as Design Option Three except that the full\_text is augmented. Spaces are added to the front of the full\_text sentence in order to get augmented Super Characters image. Each record can get multiple augmented Super Characters images until the added blanks make the sentence length equals 42 (7 words per row, max 6 rows).\label{DesignOptionFour}]{
\includegraphics[width=.5\textwidth]{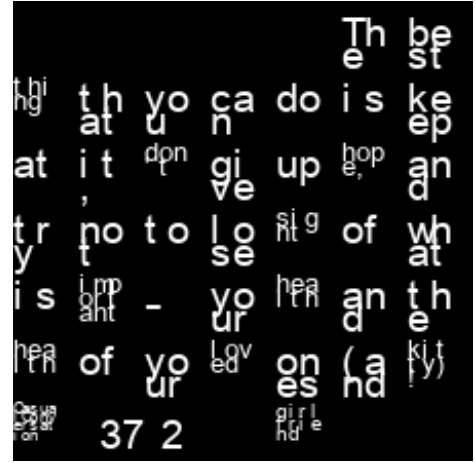}
}
\caption{Demonstrations of design options.\label{DesignOptions}}
\end{figure}

\subsubsection{Design Option One}
In this design setting, we only include the full\_text information and ignore the other attributes.
If N=7, it means each row has 7 words, and each word has (224/7)*(224/7)=32*32 pixels. In this setting we can hold up to 49 words in full\_text. For the records with words more than 49, the full\_text will ignore the words from the 49th. In this case, only 0.86\% of the training data and 1.98\% of the testing data will have to cut the sentence at 49 words. An example of this design setting is in Figure \ref{DesignOptionOne}.

\subsubsection{Design Option Two}
If N=8, it means each row has 8 words, and each word has (224/8)*(224/8)=28*28 pixels. And if we set the cutlength=40, it means that we will have 5 rows for the full\_text, and the other 3 rows will not be used for text, but all the space of the 224*(3*28) square pixels will be used for the tabular data given in the attributes other than full\_text". For the records with words more than 40, the full\_text will ignore the words from the 40th. In this case, only 2.03\% of the training data and 4.14\% of the testing data will have to cut the sentence at 40 words. We have the option to use the bottom part of the image to embed the other attributes. The id and sentenceid should be unrelated to the prediction, so these two attributes are not included. One example having the full\_text, author, wordcount, created\_utc, subreddit, score, nchar, and label is given in Figure \ref{DesignOptionTwo}.

However, the 10-fold training accuracy on this design is not good. This is partially because some of the attributes do not contribute to prediction but adds more noise instead. For example, the created time may not be very related to the prediction of the tasks but occupies a good portion of the embedding area of the image. In addition, since most of the wordcounts are centered around less than twenty, the two-dimensional embeddings of the full\_text should have better resolution if the cutlength is smaller than 40. So the font size will be larger and easier for CNN to learn.

\subsubsection{Design Option Three}
This design setting cuts the cut length of the full\_text sentence to 42, and leave the space of the last row for some important attributes, including subreddit, wordcount, score, and label. An example of this design setting is in Figure \ref{DesignOptionThree}.

\subsubsection{Design Option Four}
This is data augmentation for Design Option Three. For a small data set, we need more data with the same semantic meaning generated from the raw labeled data without adding any noise. For Super Characters, the text are projected into the image. Adding some spaces at the front should not change the semantic meaning, and at the same time increased the number of generated Super Characters images. For each sentence, if the sentence length is less than 42, we will add one space at the front and then generate the Super Characters image. This process iterates until the length of the sentence with the added space reaches 42. An example of this design setting is in Figure \ref{DesignOptionFour}.

\subsection{Experimental Results}
After comparison, only Design Option One and Design Option Four are kept for the entire 10-fold training and validation. 

For the system runs, it is limited to submit a maximum of 10 system runs. So, only the first five 10-folds models on both Design Option One and Design Option Four are tested against the 5000 testing data and submitted. The details of these 10 system runs are given in Table \ref{Table10SystemRunsEmoDisclosure}$-$\ref{Table10SystemRunsGeneralSupport}.

\begin{table}[h!]
\begin{center}
\begin{tabular}{|c|c|c|c|c|}
\hline \bf System Runs &\bf Accuracy & \bf Precision & \bf Recall & \bf F1 \\ \hline
Design Option One fold0 & 68.98\% & 33.33\% & 1.29\% & 2.48\%\\ \hline
Design Option One fold1 & 69.21\% & 33.33\% & 0.51\% & 1.01\%\\ \hline
Design Option One fold2 & 69.21\% & 33.33\% & 0.51\% & 1.01\%\\ \hline
Design Option One fold3 & 69.21\% & 33.33\% & 0.51\% & 1.01\%\\ \hline
Design Option One fold4 & 69.21\% & 33.33\% & 0.51\% & 1.01\%\\ \hline
Design Option Four fold0 & 68.98\% & 44.19\% & 4.88\% & 8.80\%\\ \hline
Design Option Four fold1 & 64.65\% & 42.99\% & 47.30\% & 45.04\%\\ \hline
Design Option Four fold2 & 70.94\% & 55.38\% & 26.48\% & 35.83\%\\ \hline
Design Option Four fold3 & 70.08\% & 51.66\% & 35.99\% & 42.42\%\\ \hline
Design Option Four fold4 & 71.34\% & 59.40\% & 20.31\% & 30.27\%\\ \hline
\end{tabular}
\end{center}
\caption{\label{Table10SystemRunsEmoDisclosure} System Run details on 10-folds validation for the task of Emotion\_disclosure.}
\end{table}

\begin{table}[h!]
\begin{center}
\begin{tabular}{|c|c|c|c|c|}
\hline \bf System Runs &\bf Accuracy & \bf Precision & \bf Recall & \bf F1 \\ \hline
Design Option One fold0 & 65.93\% & 59.21\% & 33.88\% & 43.10\%\\ \hline
Design Option One fold1 & 66.14\% & 61.84\% & 29.13\% & 39.61\%\\ \hline
Design Option One fold2 & 65.25\% & 54.65\% & 51.14\% & 52.83\%\\ \hline
Design Option One fold3 & 63.20\% & 51.13\% & 74.95\% & 60.79\%\\ \hline
Design Option One fold4 & 65.48\% & 53.63\% & 68.74\% & 60.25\%\\ \hline
Design Option Four fold0 & 67.90\% & 63.38\% & 37.19\% & 46.88\%\\ \hline
Design Option Four fold1 & 67.95\% & 64.31\% & 35.74\% & 45.95\%\\ \hline
Design Option Four fold2 & 65.80\% & 66.44\% & 20.50\% & 31.33\%\\ \hline
Design Option Four fold3 & 66.19\% & 54.61\% & 66.25\% & 59.87\%\\ \hline
Design Option Four fold4 & 66.75\% & 55.64\% & 62.32\% & 58.79\%\\ \hline
\end{tabular}
\end{center}
\caption{\label{Table10SystemRunsInfoDisclosure} System Run details on 10-folds validation for the task of Information\_disclosure.}
\end{table}

\begin{table}[h!]
\begin{center}
\begin{tabular}{|c|c|c|c|c|}
\hline \bf System Runs &\bf Accuracy & \bf Precision & \bf Recall & \bf F1 \\ \hline
Design Option One fold0 & 77.95\% & 64.18\% & 27.04\% & 38.05\%\\ \hline
Design Option One fold1 & 78.82\% & 61.83\% & 40.25\% & 48.76\%\\ \hline
Design Option One fold2 & 78.19\% & 58.10\% & 46.23\% & 51.49\%\\ \hline
Design Option One fold3 & 76.06\% & 51.62\% & 70.13\% & 59.47\%\\ \hline
Design Option One fold4 & 75.02\% & 50.12\% & 63.21\% & 55.91\%\\ \hline
Design Option Four fold0 & 78.43\% & 59.57\% & 43.08\% & 50.0\%\\ \hline
Design Option Four fold1 & 79.53\% & 62.95\% & 44.34\% & 52.03\%\\ \hline
Design Option Four fold2 & 79.13\% & 70.23\% & 28.93\% & 40.98\%\\ \hline
Design Option Four fold3 & 78.58\% & 81.94\% & 18.55\% & 30.26\%\\ \hline
Design Option Four fold4 & 78.41\% & 56.79\% & 57.86\% & 57.32\%\\ \hline
\end{tabular}
\end{center}
\caption{\label{Table10SystemRunsSupport} System Run details on 10-folds validation for the task of Support.}
\end{table}

\begin{table}[h!]
\begin{center}
\begin{tabular}{|c|c|c|c|c|}
\hline \bf System Runs &\bf Accuracy & \bf Precision & \bf Recall & \bf F1 \\ \hline
Design Option One fold0 & 73.35\% & 73.33\% & 22.22\% & 34.11\%\\ \hline
Design Option One fold1 & 72.33\% & 57.97\% & 40.40\% & 47.62\%\\ \hline
Design Option One fold2 & 72.64\% & 59.68\% & 37.37\% & 45.96\%\\ \hline
Design Option One fold3 & 72.64\% & 75.0\% & 18.18\% & 29.27\%\\ \hline
Design Option One fold4 & 73.27\% & 62.50\% & 35.35\% & 45.16\%\\ \hline
Design Option Four fold0 & 72.10\% & 61.90\% & 26.26\% & 36.88\%\\ \hline
Design Option Four fold1 & 72.64\% & 66.67\% & 24.24\% & 35.56\%\\ \hline
Design Option Four fold2 & 72.33\% & 62.79\% & 27.27\% & 38.03\%\\ \hline
Design Option Four fold3 & 75.47\% & 62.65\% & 52.53\% & 57.14\%\\ \hline
Design Option Four fold4 & 71.38\% & 56.45\% & 35.35\% & 43.48\%\\ \hline
\end{tabular}
\end{center}
\caption{\label{Table10SystemRunsEmoSupport} System Run details on 10-folds validation for the task of Emotion\_support.}
\end{table}

\begin{table}[h!]
\begin{center}
\begin{tabular}{|c|c|c|c|c|}
\hline \bf System Runs &\bf Accuracy & \bf Precision & \bf Recall & \bf F1 \\ \hline
Design Option One fold0 & 66.14\% & 55.41\% & 66.13\% & 60.29\%\\ \hline
Design Option One fold1 & 68.03\% & 61.96\% & 45.97\% & 52.78\%\\ \hline
Design Option One fold2 & 68.03\% & 57.43\% & 68.55\% & 62.5\%\\ \hline
Design Option One fold3 & 67.92\% & 72.34\% & 27.64\% & 40.0\%\\ \hline
Design Option One fold4 & 66.67\% & 66.67\% & 27.64\% & 39.08\%\\ \hline
Design Option Four fold0 & 71.16\% & 65.69\% & 54.03\% & 59.29\%\\ \hline
Design Option Four fold1 & 71.16\% & 65.69\% & 54.03\% & 69.29\%\\ \hline
Design Option Four fold2 & 68.65\% & 58.82\% & 64.52\% & 61.54\%\\ \hline
Design Option Four fold3 & 68.24\% & 66.67\% & 35.77\% & 46.56\%\\ \hline
Design Option Four fold4 & 69.18\% & 69.84\% & 35.77\% & 47.31\%\\ \hline
\end{tabular}
\end{center}
\caption{\label{Table10SystemRunsInfoSupport} System Run details on 10-folds validation for the task of Information\_support.}
\end{table}

\begin{table}[h!]
\begin{center}
\begin{tabular}{|c|c|c|c|c|}
\hline \bf System Runs &\bf Accuracy & \bf Precision & \bf Recall & \bf F1 \\ \hline
Design Option One fold0 & 78.93\% & 0.0\% & 0.0\% & 0.0\%\\ \hline
Design Option One fold1 & 79.25\% & 0.0\% & 0.0\% & 0.0\%\\ \hline
Design Option One fold2 & 73.27\% & 19.35\% & 9.09\% & 12.37\%\\ \hline
Design Option One fold3 & 77.67\% & 36.84\% & 10.61\% & 16.47\%\\ \hline
Design Option One fold4 & 79.56\% & 66.67\% & 3.03\% & 5.80\%\\ \hline
Design Option Four fold0 & 76.73\% & 16.67\% & 3.03\% & 5.13\%\\ \hline
Design Option Four fold1 & 79.25\% & 50.0\% & 1.52\% & 2.94\%\\ \hline
Design Option Four fold2 & 76.42\% & 36.36\% & 18.18\% & 24.24\%\\ \hline
Design Option Four fold3 & 80.82\% & 72.73\% & 12.12\% & 20.78\%\\ \hline
Design Option Four fold4 & 77.99\% & 25.0\% & 3.03\% & 5.41\%\\ \hline
\end{tabular}
\end{center}
\caption{\label{Table10SystemRunsGeneralSupport} System Run details on 10-folds validation for the task of General\_support.}
\end{table}

In general, Design Option Four are a little better than Design Option One, but these results are still not good. The results are a little better than constantly predict one class. 
We can see that the results on this OffMyChest data is not as good as on AffCon19 CLAFF shared task. And compared with Super Characters on Wikipedia data set, the accuracy on this data is not as accurate as well.

Several methods could be used to further improve the accuracy.
First, pretrained model may help improve. For this shared task, the size of training examples are relatively small to understand the complex definition of these 6 tasks.
Second, other data augmentation method could be introduced in order to further boost the accuracy. For example, replacing word with its synonyms.
Third, the data set is skewed data set. We can balance the data set by upsampling.

\section{Conclusion}
In this paper, we proposed modified version of Super Characters, in order to make it work on multi-modal data. In the case of this AffCon CLAFF shared task, the multi-modal data includes text data and tabular data. In addition, we deploy the models on low-power CNN chips, which proves the feasibility of applying DNN models with consideration of real-world practical concerns such as power and speed. The Super Characters method is relatively new and starts to attrack attentions for application scenarios. Pretrained models on large corpus would be very helpful for the Super Characters method, as success of pretrained model is observed for NLP models like ELMO and BERT. For fine-tuning on small datasets, data augmentation should further boost the generalization capability.

%
%
%
%

\end{document}